# State of Art of
# Adaptive Cruise Control and Stop & Go Systems

Emre Kural*, Tahsin Hacıbekir* and Bilin Aksun Güvenç*

* Istanbul Technical University, Istanbul, Turkey

*Abstract*—**This paper presents the state of the art of Adaptive Cruise Control (ACC) and Stop & Go systems as well as Intelligent Transportation Systems enhanced with inter vehicle communication. The sensors used in those systems and the level of their current technology are introduced. Simulators related to ACC and Stop & Go (S&G) systems are also surveyed and the MEKAR (Mekatronik Araştırmaları, in English "Mechatronics Research") simulator is presented. Finally future trends of ACC and Stop & Go systems and the probable advantages of them are emphasized.**

## I. INTRODUCTION

New trends in automotive technology result in new systems to produce safer and more comfortable vehicles. Many driver assistant systems are introduced by automotive manufacturers in order to prevent a possible accident either by intervening in the control of the vehicle temporarily or by warning the driver audibly and visually.

The key reason as to why researchers focus on the subject of producing safer cars is related to the statistics that exposes the serious consequences of accidents. In 2004, out of 494851 accidents, 3082 people died, 109681 people were injured and this caused approximately 0.5 billion dollars worth of financial losses in Turkey [1]. In 2003, the number of rear-end collisions were 32 286 which constitutes %54 of all accidents. In 2002, in 1023 urban accident 1685 people died and 153 billion dollars worth of damage occurred [2]. In the world, the statistics are similar: in 2002, approximately 6977 people in Germany, 7720 people in France, 3450 in UK and 6682 in Italy died due to traffic accident [2]. The statistics exposes the serious consequences of traffic accident in highways and also urban traffic.

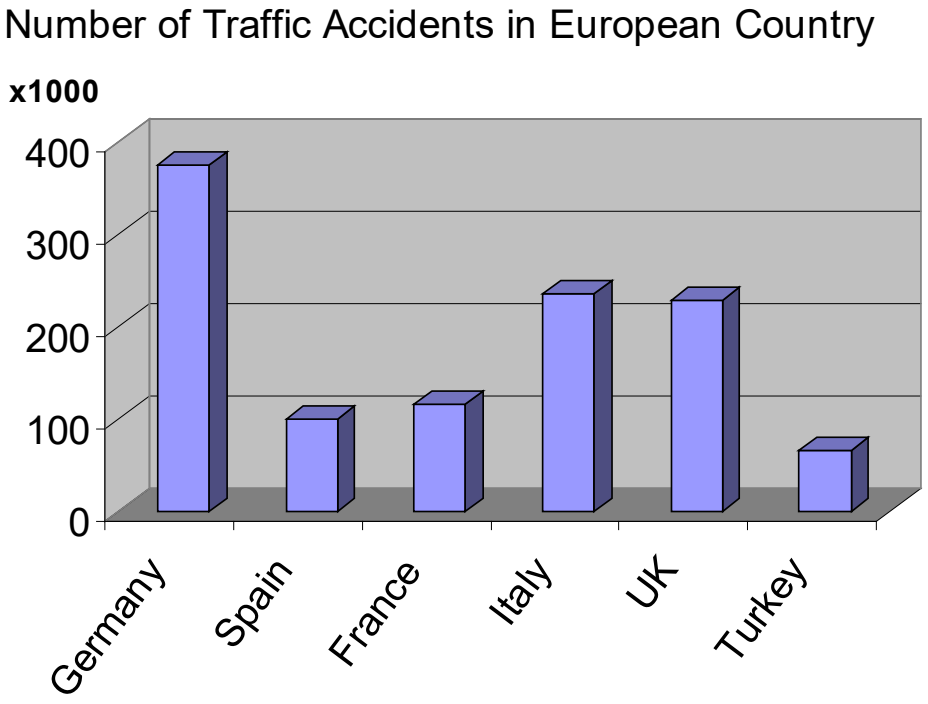

Figure 1. Traffic Accident Statistics in Europe

Consequently the automotive manufacturers take precautions which aim to reduce the severe results of accidents. To achieve this, driving assistance systems including vehicle and highway automation are introduced to decrease the number and effect of car accident due to driver error. Beside the consideration about safety, the rise in the capacity of highways, enhancement in the comfort of vehicles or fuel consumption issues push the automotive researchers to a more automated vehicles. In microscopic view an automated vehicle will provide a safer driving by compensating driver errors. And as a result the driver's routine interventions are lessened and tanks to a more controlled driving, fuel consumption will be reduced. On the other hand, the automation highways will allow the increase in the number of vehicle traveling in the roads without causing congestion [3].

Although those technology seems to be very futuristic, considerable challenges are accomplished on both academic and manufacturer levels. Actually the initiation of the studies related to automated highway or automated vehicle systems reaches to mid 1980's. In USA, Japan and Europe, programs intended to increase the capacity of highways and safety are launched by the support of the state and private foundation [4]. Furthermore, many theoretical studies is accomplished related to those subjects. Finally, up to now improvements in electronics and sensor technologies enables to realize a part of those systems and some manufacturers have already implemented those systems.

Adaptive Cruise Control (ACC) and Stop & Go (S&G) systems can be considered as the first steps of such systems that aims to assist to the driver in highways or in urban traffic. The task of them is to control the headway with the vehicle ahead along with its speed. The driver handles only by the orientation of the vehicle, so the speed and the safe distance with the forward vehicle is controlled by the system itself. The notion that separates ACC and S&G system is the speed scale in which they operate: While S&G is responsible of speed lower than 40 km/h including the stopping operation and in more congested traffic, ACC is designed for highways with higher speeds [5]. Figure 2 identifies the operation speed of S&G and ACC systems also the principal of those systems:

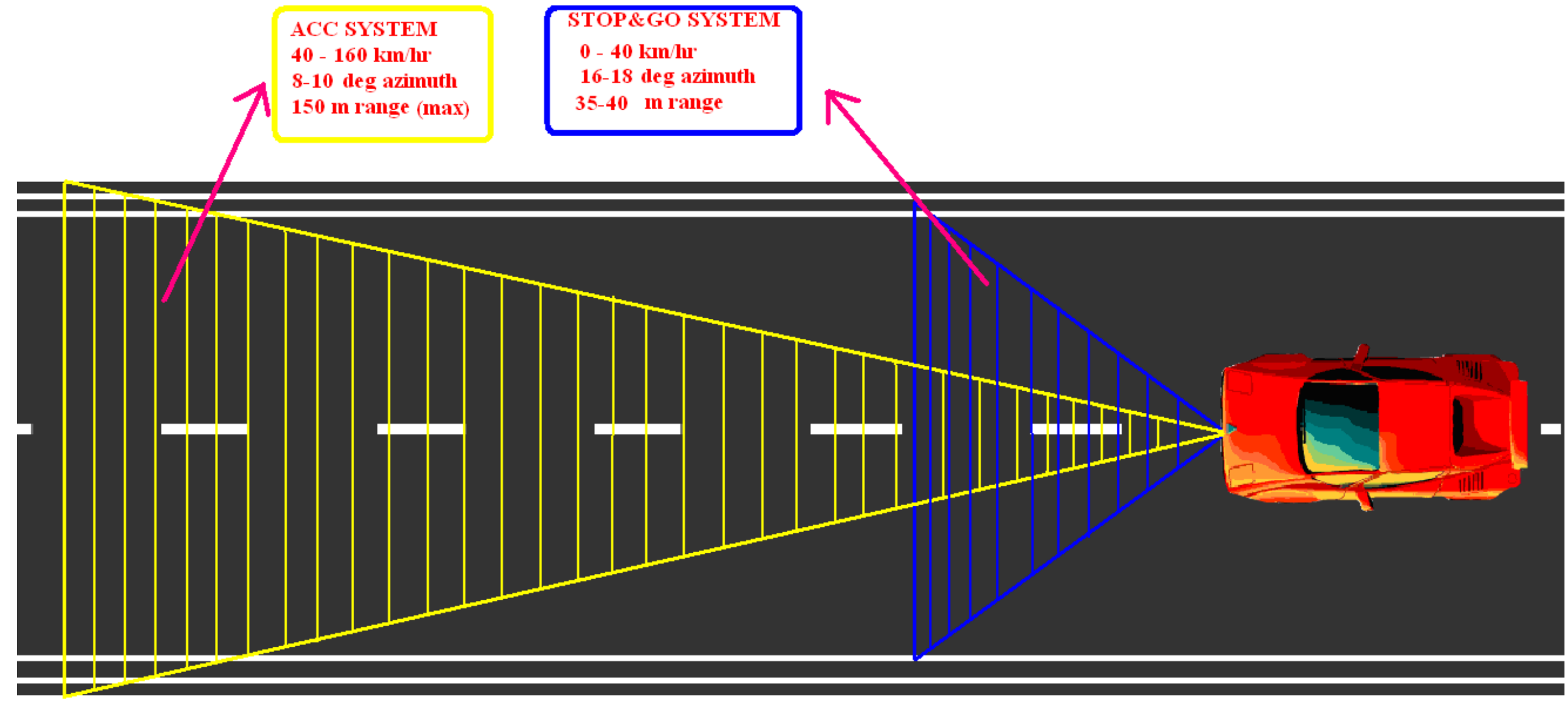

Figure 2. S&G and ACC Systems Principals

ACC and S&G systems use a forward-looking sensor to detect the presence of the target vehicle and, throttle and brake actuators to generate the required acceleration and deceleration during a following operation. Their functionality provides a safer and more comfortable drive and in the mean time alleviates the work-load of the driver. Thus the driver fatigue is lessened by the operation of these systems which lets the driver respond faster in a panic situation.

Another important issue related to ACC and S&G systems is the sensor technologies. Forward-looking sensors differ in terms of characteristic for both systems. For example, naturally S&G systems need wider angle of view, called azimuth angle, because of the more active scenarios of the urban traffic. However, ACC systems require longer safe distance due to the speed range consequently a longer detection capability is required for ACC sensors. The principal of forward-looking sensors that based on differs in today technology. Mostly radar or laser technology is used as the sensor of the present ACC and S&G systems.

Since the idea of creating an automated vehicle is examined for decades, many control theory are introduced aiming to resolve this problem. As a first step, longitudinal speed and distance problem is studied especially for ACC systems in the host vehicle point of view. Then the performance of such systems is investigated during a non uniform road condition such as cornering or lane change maneuvering. In such research works, the vehicle behavior is tried not to cause a discomfort for the driver and passengers by considering human machine interface. The designing of such systems intend to mimics the reactions of an average driver without surprising when it operates [6], [7].

Even though the ACC designs focus on the driver's safety and comfort, there are considerable studies on its effects on different issues such as traffic flow and the environment. Also, some automated systems aimed at reducing the traffic intensity and this can be seen as a possible solution to the traffic flow problem. The platoon systems which can be considered as a future generation of ACC or S&G systems are introduced to reduce the congestions in highways. Platoon is a set of vehicle traveling together with a specified distance. The communication protocols or spacing policy differs with the design. However the common objective is to form of a platoon with shorter space than usual drivers can not which will increase the capacity of highways. Furthermore, since the relative speed is lower with respect to usual case in case of an accident the impact energy is lower. Another advantage that a platoon provides is a less aerodynamic drag for the vehicle preceding the first one which is related directly to the fuel consumption [8]. ACC systems effect on fuel consumption is also an ongoing study in academic environments by means of a smoother cruise [3].

Simulator studies are also an important issue for ACC or S&G systems. They constitute a milestone in the design of such system. On one hand, a virtual environment reduces very high costs of road tests and on the other hand it provides a safe location for tuning the system parameters and to test real human behavior in a real traffic situation. There are many ACC simulator designed on this purpose. One of them is settled in MEKAR, as a low cost PC based real time simulator. MEKAR is a research laboratory, founded in 1997 in the department of mechanical engineering of Istanbul Technical University, which is an acronym of Mechatronics Research (Mekatronik Araştırmaları) and Mechatronic Vehicle (Mekatronik Araç).

This paper reviews the current state of the art of ACC and S&G systems investigate the last innovations in those systems, the continuing academic studies and also survey the studies on Automated Highway Systems briefly. The system description and the control problems will be reviewed. The sensor requirements of the system and sensor technology are outlined.

The rest of the paper is as follows: In part 2, a system description will be given by presenting first the big projects throughout the world about these subjects and then ACC, Stop&Go and Platoon systems will be studied in detail with a literature review. In part 3 the sensor requirement will be mentioned. And in the following part inter vehicle communication related studies will be presented. Finally, the paper concludes with the overview of the ACC related simulator studies including MEKAR ACC simulator and the future aspects of ACC and Stop & Go systems.

## II. DEVELOPMENT STAGES OF INTELLIGENT VEHICLE SYSTEMS, ACC AND STOP&G O

The early stages of Intelligent Vehicle Highway Systems (IVHS) are initiated with the efforts of programs supported by the governments and automakers. Especially in Europe, USA and Japan those programs founded with considerable budgets, aim to produce the vehicular automation with futuristic approaches [3].

EURAKA which is a unique institution, conducted many projects executed by automotive industry. One of them was called PROMETHEUS (Programme for a European Traffic system with Highest Efficiency and Unprecedented Safety) which introduces feasible technology to IVHS. This program of the European motor industry began as a cooperative research and development effort in the mid 1980's. The PROMETHEUS research program united car companies from six countries with automotive suppliers, electronics companies, government research laboratories and universities. The member bodies began work in 1986 on pre-competitive research exploring ways of achieving safer driving, smoother traffic flow and improved travel and transport management.

In the United States, the U.S. Department of Transportation's Intelligent Transportation Systems (ITS) Joint Program Office and the Partners for Advanced Transit and Highways (PATH) program founded by California Department of Transportation and Institute of Transportation Studies of the University of California at Berkeley study advanced vehicle and highway systems. The program aimed the development of a full automatic vehicle system able to communicate with other vehicles and travel along a platoon in the highway by using control algorithms in both lateral and longitudinal directions. The PATH project was not only dedicated to Intelligent Cruise Control systems but also some automated systems for heavy trucks, transit-buses and stop and go applications, immobile object or pedestrian recognizing systems etc. [11].

Lastly, Japan's Advanced Cruise-Assist Highway System Research Association studies on the systems which makes driving a completely automated experience. The Advanced Cruise Assist Highway System (AHS) is one of the most advanced systems in the ITS field. The goal of AHS is to reduce traffic accidents, enhance safety, improve transportation efficiency as well as reduce the operational work of drivers. A number of related effects are also expected. In Japan, AHS research is being carried out in the following fields: AHS-"i" (information): focusing on providing information; AHS-"c" (control): vehicle control assistance; AHS-"a" (automated cruise): fully automated driving [4].

The experimental result of the studies related to AHS gave satisfactory results and the works on these subjects continue to create the next steps of automated vehicles. However due to the economical difficulties or feasibility problems AHS studies are replaced by more practical solutions in Intelligent Vehicle Systems such as ACC or S&G and other collision avoidance systems. ACC and S&G systems are currently being developed by many automotive manufacturers, private and combined research groups around the world. The most important advantage of the ACC systems that they do not require a supplement infrastructure such as an automated highway or automation in other vehicles. Furthermore, the knowledge and experience gained from the national programs can straightforwardly convert into the application of those systems.

In order to focus more into these systems and expose the difference each system will be observed separately.

### A. *Adaptive Cruise Control*

A longitudinal direction driver assistance system, ACC, attenuates the driver's routine by assisting with the operation of brakes and the accelerator. It uses a sensor in order to measure the relative speed and relative distance between the host vehicle on which the ACC system is installed and the target vehicle in front. It automatically adjusts the vehicle speed to maintain the distance set by the driver and when there is not a target vehicle, it accelerates until the pre-set-speed, in cruise control mode. Electronically controlled vacuum booster controls the brake system in order to provide the desired deceleration during a following scenario [12]. Similarly throttle actuation allows a possible acceleration if it is necessary. However, the authority of these actuations is limited due to the safety reasons [13]. The forward looking sensor system senses the target vehicle ahead and the speed of the host vehicle is controlled according to this target vehicle by insuring a safe distance related to the time gap entered by the driver. The speed of the vehicle is tracked according to the preceding vehicle in such a way that it does not excess the pre-set cruise control speed. This speed is used once there is not a target car or the preceding vehicle increases its speed over this value.

Although ACC is launched as a comfort system that substitutes the driver in controlling headway distance and speed during a following condition, the system introduce a safe cruise due to a controlled acceleration and deceleration behavior [3]. Actually driving skill, driving habits, and response time of the drivers are so various because of the differences between age, gender, healthy and psychological condition etc. An inappropriate or late response of the driver to a change in traffic flow will cause fatal accidents and consequently financial problems. On the other hand, ACC can respond much more quickly and precisely than human drivers can to any change in speed. A vehicle using adaptive cruise control typically brakes sooner and more smoothly than one without the system.

Many control schemes are accomplished related to longitudinal behavior of ACC in highways [14],[15],[16]. However, the performance of the system during a not-uniform road condition such as cornering maneuvers should be carry out carefully. Tracking of the target vehicle during a curve among the other vehicles in adjacent lanes is required. To overcome this problem a prediction algorithm based on the sensor data obtained from the yaw rate and/or steering angle is used [17], [18].

When control studies are investigated systematically there are three types of control approach to solve the problem. Firstly, the longitudinal control where the assumption of the vehicle path is completely straight, and the control solves the two vehicle spacing problem. The interaction of the vehicles on longitudinal dynamics and on the host vehicle point of view. The studies generally take into account the problem with two vehicles and for the simulation a simple, reduced order longitudinal vehicle dynamics including engine and brake dynamics are studied. Microscopic traffic flow analysis is done and the lateral interaction of vehicles is omitted [19],[20],[21]. The response of the host vehicle to the changes of speed of the preceding one is examined during acceleration, deceleration or go back to the cruise control mode. For the second approach, in addition to the longitudinal the lateral behavior of the system is investigated, and the scenarios are expanded such as lane changes or following during cornering [18]. To achieve this, lateral dynamics of vehicles are added to accomplish the required maneuvers [22]. And lastly, the platoon approach are studied as a result of different programs such as PATH or PROMETHEUS, where the situation of following of more than two vehicles are investigated during different scenarios [23],[24]. The system is observed in multi car simulations, and the traffic flows and dynamics are modeled in a more general sense. The macroscopic traffic characteristics is also observed [25],[26]. Especially the studies concerning the string stability problem of a platoon examined [10]. The string stability problem of a vehicle platoon has been studied since the late 1970's. The term "string stability" refers to the non-amplifying upstream propagation of vehicle speed perturbation through a string of vehicles.

The control schemes, are generally constituted of two hierarchical structure [27],[19]. Firstly, ACC control is responsible for computing the required acceleration or deceleration of the vehicle according to a control algorithm. This upper level algorithm produces the required control input for either brake or throttle actuators by considering comfort and safety issues. The sensor inputs are directly involved in this control block and relative distance and relative speed constitutes the inputs for the controller. Also, the multi-target tracking among different objects and path estimations can be considered as a task of this section. On the other hand, the control inputs calculated are transferred to the actuators and the control issues are realized by the low-level controllers assuring the actuation of the correct outputs based on the desired inputs.

In observing the effect of ACC systems on traffic flow, distinct control approaches are considered. Those approaches are based on;

- Constant Space Headway Control: In this approach the distance between vehicles is assumed constant [28].
- Constant Time Headway Control (CTH): In this approach, the control input is a function of the time gap (which can be selected by the driver).And when it is multiplied by the vehicle speed constitutes the controllers desired distance. So the system's string stability is guaranteed while the smooth acceleration/deceleration is ensured.[29]
- Variable Time Headway Control (VTH): In this strategy the desired distance is regulated according to the relative velocity of the vehicles. In this case, the 'time gap' value is changing continuously [30].

The control problem can be summarized such that the relative speed and relative distance is forced towards the phase plane origin. In Özgüner et al. [19], the problem is solved by proposing different control actions defined in the phase plane by different region called decision regions on which the driver behavior is emulated as close as possible. They aim to design a hybrid controller for the longitudinal control of vehicles in highway which switches between different control actions. The distance and the relative velocity is forced towards the origin of the phase plane smoothly and the high frequency switching in the decision points and at the origin of the phase plane is avoided. A supervisory hybrid controller switches between smooth-constant acceleration / deceleration and linear control region and the study is presented with experimental and theoretical results.

There have also been also some PID control applications in the development of ACC studies: In Ioannou et al. [31] a PID controller based on constant time headway policy is proposed and proved to provide both local and platoon stability where it is theoretically robust against the sensor noise delays.

In Yanakiev and Kanellakopoulos [32], two control approaches (PID with quadratic derivative term and adaptive PID) were compared. Hedrick and Yun have designed two level controllers for heavy-duty truck vehicles where the control problems differ from passenger cars due to the different structure of these vehicles: A low level controller handles the brake period in brake systems and fuel injection period in a turbo-charged diesel engine [33]. And the higher level controller is based on a sliding mode control structure.

Gain scheduling and adaptive controllers are widely used in the solution of the control problem. In [29], an adaptive PID throttle controller is proposed and compared to constant term PID and adaptive controllers. The paper designs control architecture in the longitudinal dynamics. At first the constant time headway based control objective is traced. Then the sub-controllers are presented for throttle and brake. For throttle, a PID with fixed gain and gain scheduling algorithm is introduced. Also alternatively an adaptive controller is presented. A discontinuous braking controller is given. Finally, the switching algorithm between throttle and brakes is introduced.

As a part of the PATH project Hedrick's studies related to ACC or more generally platoon control is based on sliding mode control architecture. As an extension of ACC, Cooperative Adaptive Cruise Control (CACC) is introduced. This system provides an inter vehicle communication among numerous vehicles, and it constitutes a hybrid control structure [27],[34]. CACC allows communication of the host vehicle with the lead one. It is claimed that CACC system provides closer following distance in contrary to ACC system, because of the information provided by the reliable and quicker communication [35]. Furthermore, the communication supplies the maximum braking capability or braking rate of the lead vehicle, which will warn the driver of the host vehicle sooner with respect to a traditional Forward Collision Warning System. In the light of this information, the system is separated into seven different modes where different systems (ACC, CACC, Cooperative Collision Warning (CCW) etc.) are switched according to the scenarios by the supervisory controllers [27]. Actually, the structure of the control architecture can be scaled into three layers: The most basic controllers deal with actuator and basically continuous control algorithm are selected. The maneuver coordination layer is a hierarchical controller based on discrete states, each of them send desired input signal to the basic controllers. And finally the supervisory control achieves coordinated maneuvers between vehicles. Practical implementation of the study is executed as a part of the PATH project.

The focus of [35] is on direct (no base-station) vehicle-to-vehicle wireless communication at 1 to 5 km distances, which is studied at the Ohio State University (OSU). The classification of information, effect of information transmittal, physical layer, MAC protocols for delivery of real-time information and output of simulation the ES-DCF and DB-DCF protocols are discussed in [35]. Furthermore, In [36], different delay time equations with different cases of the inter-vehicle communication are examined and also compared with each other to evaluate the effect of communication on rear-end collision avoidance. A unified model for delay time computing equation are obtained and simulated to present validation of the general delay time equation by Liu and Özgüner.

Yi et al. proposed a throttle/brake control law that generates the desired control input and supported their theory with experimental results [15].

There is insufficient number of studies concerning system lateral performance during cornering scenarios. Most detailed studies are presented here briefly. In [16] a target identification method using yaw rate is proposed. The method is based on the phase chart between the lateral component of the relative velocity and azimuth of a preceding vehicle. In a similar study of one of the paper, two new methods called Method Curve and Method Lane are proposed [17]. A phase plane constituted by azimuth angle and relative velocity defines the curved entry of a preceding vehicle or lane change scenarios of the vehicles. In [18] the target vehicle lateral position estimated with respect to the host vehicle's projected path.

Most of the theoretical work related to longitudinal control is on the driver comfort and safety from the host vehicle point of view. However, in many publications the study related to traffic stability, highway traffic flow dynamics and highway platoon driving issues are observed. The string stability concerning ACC algorithms is widely studied in the literature. It is known that the string stability problem can be handled if the constant time headway policy is used. However a system that enables the intercommunication allows a string stable platoon with constant space [36]. On the other hand, Yanakiev et al. claims that constant time headway results with larger spacings which elongates the length of the platoon and influences the traffic capacity. So they introduces variable time headway policy [37]. In [10,38] Peng and Liang proposed a string stable optimal control algorithm and verified their result on a traffic simulator. In [39], the performance of ACC is compared to a manual driving vehicle, in terms of string performance using a human driver model.

Raza et al [40] tried to design an overall system that covers different driving environment. The system called Vehicle Longitudinal Control System (VLCS) and it includes different modes such as autonomous vehicles, cooperative vehicle following and platooning. The control system of the vehicle is built on supervisory controller and throttle/brake controller. Moreover, supervisory controller decides operation modes of vehicle and also evaluates the inputs which come from the driver, infrastructure of roadside beacon, on board sensors and other vehicles.

The first mode of the VLCS is Intelligent Cruise Control (ICC) that allow automatic vehicle following with supervision of the driver. Second mode: Cooperative Driving without communication between vehicles is obtained by adding the roadway to vehicle communication capability that mandates the speed another mode Cooperative Driving with communication between vehicles allows the vehicle to coordinate maneuvers and exchange the information about road conditions, specifications of vehicle. Lastly, Platooning mode is described as vehicles communicating with each other and the roadway to coordinate themselves on the way.

In another interesting study, Zhang and et al. [41] concentrated on designing an autonomous intelligent cruise control (AICC) which uses information from the vehicles ahead and also behind to guarantee individual vehicle stability as well as platoon stability in both directions under the constant spacing safety policy. Moreover, AICC law is able to guarantee platoon stability for a speed dependent desired spacing at all speeds. Zhang et al. give information about the problems and different approaches of platoon stability and also present that only platoon stability does not provide the vehicle stability. The controller must be designed to guarantee both vehicle and platoon stability. The model and controller are tested by forming a platoon that contains leader and four following vehicle in different cases. The effects of the AICC is observed on traffic flow and also compared with three human driver models proposed in the literature by Ioannou et al. [42]. The constant time headway approach is used to eliminate slinky effect with appropriate control system and also the performance of the AICC is compared with these three drivers. Results of the simulations show that AICC provides large traffic flow rates and smoother traffic flows by eliminating the human delays, large reaction times and obtaining shorter inter vehicle safety space.

Intelligent Cruise Control (ICC) laws for passenger vehicles are presented in [43]. Kyongsu et al. mentioned about the types of the longitudinal control and also low performance of the linear PID-controller because of noisy sensor data. Numerical values of the ICC system components were presented and also control algorithms and equations of the system components are given in [43]. LeBlanc et al. presented the results of the Crewman's Associate for Path Control (CAPC) study that is an automatic road-departure warning system for motor vehicle drivers. The main difference of the CAPC from the autonomous road or vehicle following system is automatic assistance maintained only until the driver recovers control or until the vehicle has been brought to rest by a co-driver.

The most researched area in ACC systems is the control methodology. Different and interesting studies are summarized below. Hatipoğlu et al [44] explain development of a supervisory hybrid controller which is designed to switch between different controls actions to obtain a smooth intelligent cruise control (ICC) structure. Different decision regions are defined on the phase plane and also three different control actions are studied. Controllers were connected to each other to mimic the behavior of an experienced driver by using hybrid model and Kalman filter which is designed for state observation. Longitudinal and lateral control algorithms were developed for the radar based convoying problem for truck-trailers under the assumption that the convoy consists of only two trucks by Haskara et al [45]. The vehicle following algorithm and the evaluation logic of the sensory information is presented to find the desired speed for longitudinal control in [45].

Acarman et al [46] presented a new driver assistance system that contains automated Control Authority Transition (CAT) system and its modules for planning, decision making, execution, reference or expert driver and sensor modules. CAT system is a concept of transferring some of the driver's control authority to an automated system that has no new safety problems such as distraction and information overload [46]. In [46], simulation result of some accident scenarios related to some modes of operation of the CAT system is given to show its validity to prevent accidents.

### B. *Stop & Go*

Stop and go cruise control is an extension to ACC which is able to automatically accelerate and decelerate the vehicle in city traffic and Stop and Go situations. It reduces the driver workload in suburban areas, where ACC systems are practically ineffective.

Due to the more complex driving environment and more rigid sensory requirements in lower speeds, the challenges in developing stop and go systems are more than those of ACC systems. The sensory requirement for the driving environment detection is much more extended for a stop and go cruise control system [3]. The sensor requirements differ from ACC applications. First of all, the field of view of the host car is expanded. Furthermore, the control structure changes because the speeds range changes. More aggressive acceleration and deceleration limits are introduced [47].

There are some control problems of the Stop & Go which are changing the limitation of acceleration and deceleration in such a way that does not cause discomfort for the driver when the speed range decreases and also the sensor resolution that should be high enough to handle the small distance and relative velocities.

Venhovens et al [38] traces the borders of the S&G tasks. It mentions that the system should be performed in a congested traffic environment where the driver is dependent on traffic flow. There are stopped and discontinuous starts. Hedrick et al [49] have proposed the studies of stop & go application in PATH project in Berkeley. The control structure resembles the usual ACC structure and a desired distance is given with an empiric expression based on the human behavior in [39]. In [39], after upper level controller the lover level throttle and brake controllers are introduced, the lower level controller tries to produce the required acceleration that the upper level desires. In this point of view the inverse of the look-up table is used, also the desired brake torque is ensured by controlling the master cylinder's pressure. Furthermore, driver behaviors are another important issue of the stop & go. Canale and Malan [50] have studied designing the ACC system which has ability to operate at stop & go situations in urban traffic, in such a way that it represents an average driver behavior to accomplish that the system is tuned by the data collected with 20 different drivers. Yi et al [51] have presented a vehicle speed and vehicle-to-vehicle distance control algorithm for a vehicle S&G system whose control algorithm consists of speed and distance control algorithms and a combined throttle hake control law. Linear Quadratic (LQ) optimal control theory has been used to design vehicle desired acceleration for vehicle-to vehicle distance control [51].

## III. SENSOR DESCRIPTION

ACC and S&G systems' most critical requirement that will provide an adequate control without intervention of the driver is the sensor systems that are responsible to supply the correct data between the host and preceding vehicle. The controller uses this relative speed and relative distance data and computes the required control input to the actuators of vehicles. In today's technology there are mainly two types of sensor system: Lidar and radar based based forward looking sensors.

### A. *Lidar Based Sensor:*

Lidar Based sensors use the light detection and ranging technology. The lens of the sensor optically disperses the laser beam into adjacent regions and transmits them. The coverage of the beams does not overlap and there is not a separation as well. The most important advantage of the laser beams is their ability in multi-tracking. For example, Delphi's Forewarn® ACC-L Lidar System, which has 12 adjacent laser beams, can identify 6 objects within each region [5]. The sensor uses a pulse signal in order to measure the distance of the target. The Lidar products make use of light beams and are therefore environment sensitive. The weather condition affects the reliability of the measurement negatively [52].

### B. *Radar Based Sensor:*

Another type of forward looking sensor is microwave radar. The radar configuration used in ACC and S&G systems differs in terms of frequency and specification. ACC system requires long range radar due to the high speeds and longer detection distance. Long range radar devices use FMCW with an indirect flight-time measurement. A CW radar compares the frequency of the transmitted signal with the frequency of the received signal. Transmitted frequency is compared to the received and Doppler shifted signal and the flight time of the signal is computed resulting in the distance. The frequency of ACC radar is 76-77 GHz and the detection distance is 150m with 8° of field angle so called azimuth. However, S&G systems need a wider field of vision due to the traffic characteristics in urban area. The azimuth angle of short range radar used in S&G systems has an azimuth angle of 16°, with a detection distance of 40m. The frequency used in this type of sensor is 24-25 GHz. The following figure compares the characteristics of ACC and S&G radars.

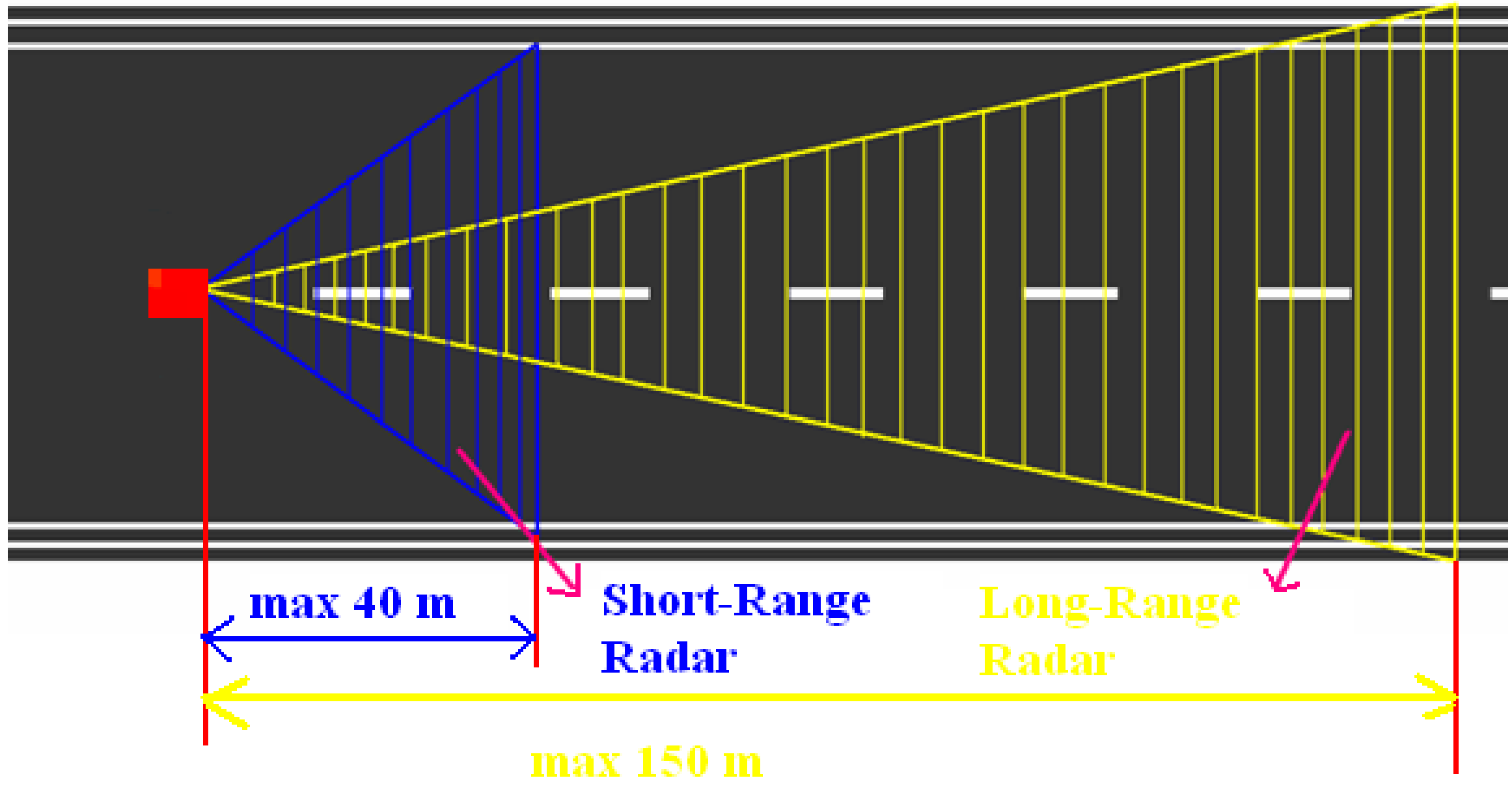

Figure 3. Characteristics of Long and Short Range Radars

There are various antenna type for sensor systems such as monopulse, single scanning beam and switched beam. The latter is an electronically sweeping device. The monopulse is actually three antenna systems where one of them transmits and the two others receive. This configuration is able to identify the azimuth angle of the target object also has the capacity of object multi-tracking [53].

The development of inter-vehicle communications systems is one of the most studied subjects in ITS industry. Communications between vehicles would present a number of potential safety benefits. The principle behind the system is that one vehicle is able to transmit speed, road and warning information to other vehicles on the road. This form of system may be implemented strictly between vehicles or through a roadside repeater. The major potential benefit situations include accidents or road obstacles where following vehicles can be alerted.

The difficulty with such a system is that it will clearly only function in equipped vehicles only. Therefore, it would not be effective without a high level of usage. It is unlikely that a purchaser would select an option that is dependent upon purchases by other motorists. Therefore, this system can only be seen as a possibility in the fairly distant future.

Recent advances in wireless data communications technologies have led to the development of Vehicle Safety Communication (VSC) applications. This new breed of automotive technologies combine intelligent on-board processing systems with wireless communications for real-time transmission and processing of relevant safety data to provide warnings of hazards, predict dangerous scenarios, and to help avoid collisions. VSC applications rely on the creation of autonomous, self-organizing wireless communication networks, called ad-hoc networks, connecting vehicles with the roadside infrastructure and with each other. While the technical standards and communication protocols for VSC technologies are still being developed, it becomes vital to consider potential value and ethical implications of the design of these new information technologies [56].

On the other hand, an interesting study by Farkas et al [57] explains development of forward-looking chirp monopulse radar which senses echoes from a highway stripe frequency selective surface (FSS) to provide lateral position information. Moreover, it senses echoes from vehicles and other obstacles in the lane ahead to provide distance information for automated cruise control.

## IV. ACC SIMULATORS

There exist different types of vehicle simulators built for different purposes. While some of them are constructed with quite wide resources and especially for general purposes inside big facilities, others are less expensive and assembled with moderate budgets and focus on more specific research projects. Also, there are many companies specializing on manufacturing driving simulators for commercial purposes. This constitutes a part of the research projects of big vehicle companies such as Volvo, Renault or Ford. One of the important experiments carried on the simulator was the development of the ACC algorithms. The objective of the development of the ACC on a simulator is to tune and evaluate the system in a virtual environment instead of the real highway traffic condition which may be risky for both the host and the other vehicles. In addition, the subjective comfort assessment is easily done on such a non-fixed simulator.

As mentioned above there are important simulator projects built for more general purposes. One of them is NADS which is claimed to be the “most advanced driving simulator in the world”. The simulator was completed in 2001 after 9 years in the University of Iowa and was dedicated to research on highway safety and vehicle system design [58].

Another big project dedicated to vehicle dynamics simulation is TNO’s VEHIL (Vehicle Hardware In the Loop) built in 2003 in Netherlands. The facility is founded for the research of intelligent vehicle systems such as ACC, stop&go, platooning, precrash sensing etc. The test environment differs from a classical driving simulator where human intervention is indispensable. VEHIL offers the possibility of handling the expensive road tests of intelligent vehicle systems in a laboratory condition. Subjecting the sensor and actuator of a vehicle to the real driving conditions, the functionality of intelligent vehicle system is evaluated.

A full intelligent vehicle is placed on a chassis dynamometer where each wheel is supported by a drum. The vehicle accelerates or decelerates over this platform up to 250 km/h. The rest of the test system constituting the other vehicles is simulated with "highly dynamic automatic guided vehicles". These four wheel mobile platforms are able to represent all longitudinal, lateral and yaw dynamics of a vehicle. Finally, a visualization system completes the realistic behavior of the simulation system.

Renault's dynamic simulator was installed in 1999 at the Research Department of Renault dedicated to automotive research. The simulator consists of a platform of 6-DOF, a vehicle cabin and the complementary audiovisual systems. Also, the dynamic force feedback is provided on the steering wheel, accelerator clutch and brake pedals. Thanks to the moving platform the translational and rotational dynamic behavior can be simulated. The simulation software SCANeR II was developed by the Research Department of Renault [60].

In MEKAR since 2004 a PC based low cost ACC simulator is being developed. Two game type Steering are used in Wheel, driver- in the loop simulation. A 5 dof vehicle model including ACC controller and a 5 dof target vehicle allow on ACC algorithm development environment in real time simulations with xPC Target and Simulink models. A static drivetrain model and an engine model as a look-up table are used by the two vehicles models. xPC target allows to the communication of two computer. The real time code generated and downloaded to the host PC. And in the host PC the user inputs and visualization are realized. The user inputs are included to the simulations thanks to the two game type steering wheel attached to host PC with USB port. A visual environment is created using MATLAB/Virtual Reality Toolbox. Two levels ACC control algorithm is realized using a linear control scheme and a state driven scheme where the control inputs are saturated in order to prevent uncomfortable behavior of the ACC vehicle. Low levels controllers generate the required acceleration and deceleration and high level controller identifies the scenario happening and evaluates the inputs of the low level controller inputs. Many scenarios such as lane changes or cut-in scenarios can be identified by the high level controller. The control action is converted to the torque of the engine by an inverse engine model. As a result MEKAR simulator and its model allow to the online simulation with two users and offline simulations without user interface. MEKAR simulator is shown below as Figure 4. Furthermore, a typical simulation result and the view of the simulator are seen in the Figure 5.

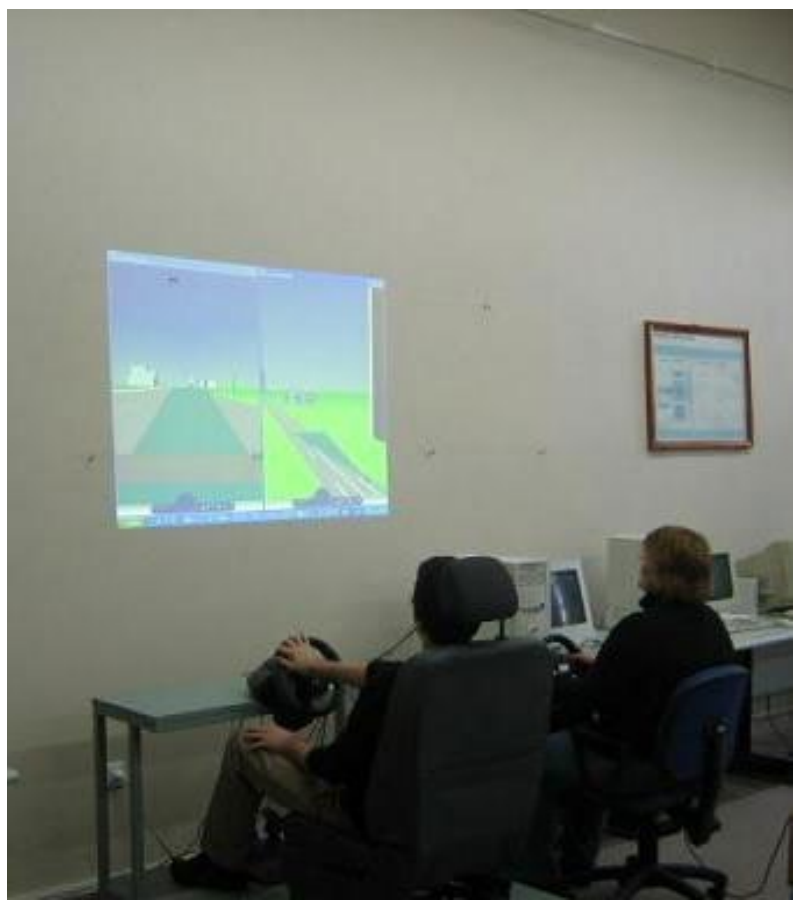

Figure 4. MEKAR Simulator

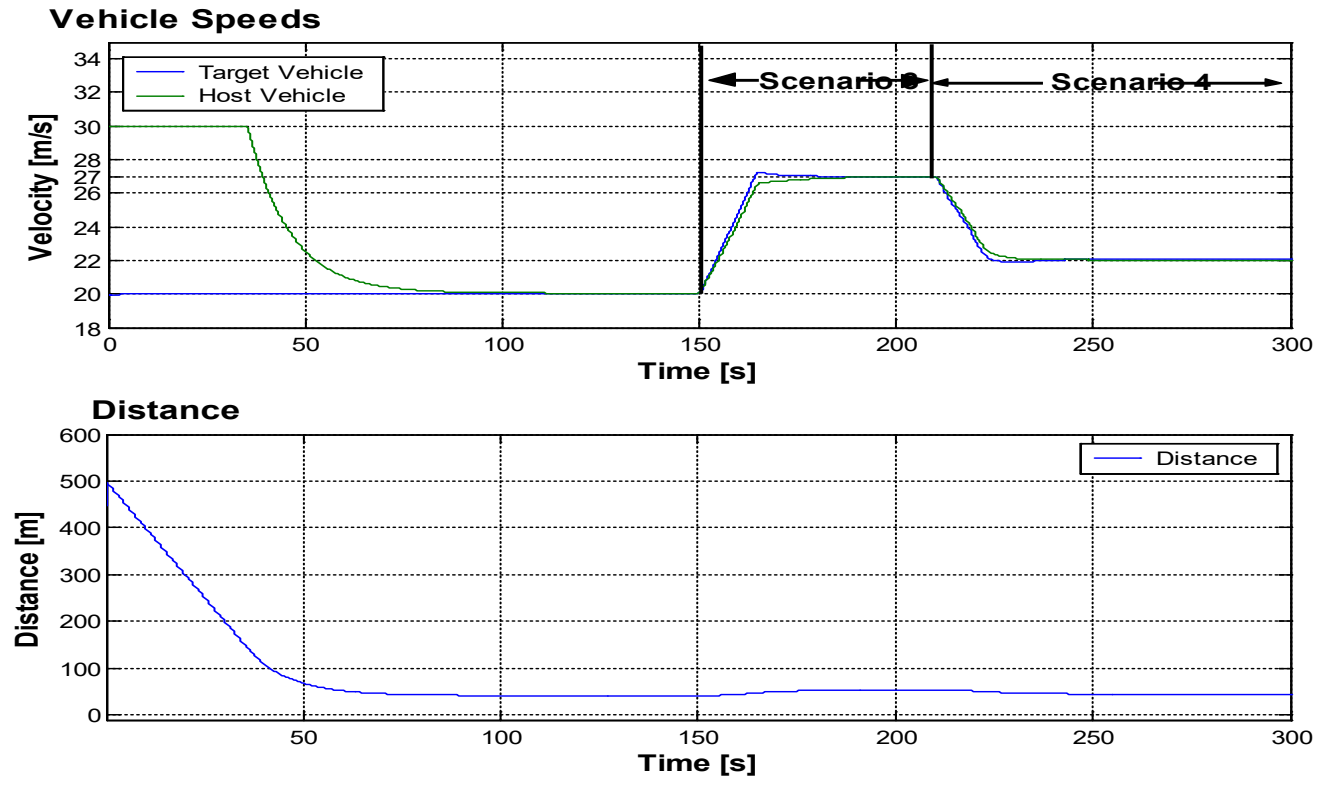

Figure 5. Simulation Result for ACC operation during following, target accelerating and decelerating scenarios

## V. FUTURE TRENDS AND MORE RECENT DEVELOPMENTS

In this study, the current state of the art of ACC, S&G and Platoon systems has been reviewed in detail. Sensor requirements, inter-vehicle communication studies and overview of the ACC related simulator studies including the MEKAR ACC simulator have been presented. Besides the intense theoretical and practical studies, the sensory problems should be carried out in order to implement driver assistance systems in the future cars. Also, the complex scenarios including cornering, lateral detection or path estimation should be identified by the system. Especially the performance of those assistance systems in urban traffic condition should be analyzed carefully. Another important aspect is the studies that involve the human factors with these semi-automated driving control systems. There are some studies including human-centered approaches for designing, describing and evaluating these driving assistance systems.

There have been many developments that were reported in the literature since the original writing of this paper in 2007. Based on developments in Vehicle to Vehicle (V2V) communication, significant work has been done on cooperative car following using Cooperative Adaptive Cruise Control (Wang et al; Ma et al, 2021; Wang et al, 2020; Ma et al, 2020; Yang et al, 2020; Emirler at al, 2018; Tamilarasan and Guvenc, 2017; Guvenc et al, 2012; Tamilarasan et al, 2017; Emirler et al, 2013; Hartavi et al, 2012; Uygan et al, 2011). Model Predictive Control (Emekli and Aksun-Guvenc, 2016) has been applied to ACC (Kural and Aksun-Guvenc (2010) and later to CACC. The model regulator also called the disturbance observer can be used to regulating the ACC or CACC plant model about its nominal one in the presence of model uncertainty (Aksun-Guvenc et al, 2003; Aksun-Guvenc and Guvenc, 2002; Guvenc and Srinivasan, 1994; Guvenc and Srinivasan, 1995). Parameter space based robust control methods including their frequency domain extension are also very successful in treating uncertain plants when they are used together with the model regulator approach (Guvenc et al, 2017; Aksun-Guvenc et al, 2017; Aksun-Guvenc and Guvenc, 2001; Aksun-Guvenc and Guvenc, 2002; Emirler et al, 2015; Orun et al, 2009; Demirel and Guvenc, 2010; Oncu et al, 2007; Guvenc, et al, 2021; Wang et al, 2018).

ACKNOWLEDGMENT

The authors acknowledge the support of the European Union Framework Programme 6 through the AUTOCOM SSA project (INCO Project No: 16426) in this paper.